\documentclass{article}

\usepackage{arxiv}

\usepackage[utf8]{inputenc} 
\usepackage[T1]{fontenc}    
\usepackage{hyperref}       
\usepackage{url}            
\usepackage{booktabs}       
\usepackage{amsfonts} 
\usepackage{amsmath}
\usepackage{nicefrac}       
\usepackage{microtype}      
\usepackage{cleveref}       
\usepackage{lipsum}         
\usepackage{graphicx}
\usepackage{natbib}
\usepackage{doi}

\title{Six Sigma for Neural Networks: Taguchi-Based Optimization}

\date{}

\author{ \href{https://orcid.org/0009-0002-3124-633X}{\includegraphics[scale=0.06]{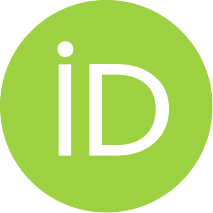}\hspace{1mm}Sai Varun Kodathala} \\
	Research and Development\\
	Sports Vision, Inc.\\
	Minnetonka, MN \\
	\texttt{varun@sportsvision.ai} \\
}

\renewcommand{\headeright}{Technical Report}
\renewcommand{\undertitle}{Technical Report}

\hypersetup{
	pdftitle={FAST OTSU THRESHOLDING USING BISECTION METHOD}
	pdfsubject={q-bio.NC, q-bio.QM},
	pdfauthor={Sai Varun, Kodathala},
	pdfkeywords={First keyword, Second keyword, More},
}

\begin{document}
	\maketitle

\begin{abstract}
	The optimization of hyperparameters in convolutional neural networks (CNNs) remains a challenging and computationally expensive process, often requiring extensive trial-and-error approaches or exhaustive grid searches. This study introduces the application of Taguchi Design of Experiments methodology, a statistical optimization technique traditionally used in quality engineering, to systematically optimize CNN hyperparameters for professional boxing action recognition. Using an $L_{12}(2^{11})$ orthogonal array, eight hyperparameters including image size, color mode, activation function, learning rate, rescaling, shuffling, vertical flip, and horizontal flip were systematically evaluated across twelve experimental configurations. To address the multi-objective nature of machine learning optimization, five different approaches were developed to simultaneously optimize training accuracy, validation accuracy, training loss, and validation loss using Signal-to-Noise ratio analysis. The study employed a novel logarithmic scaling technique to unify conflicting metrics and enable comprehensive multi-quality assessment within the Taguchi framework. Results demonstrate that Approach 3, combining weighted accuracy metrics with logarithmically transformed loss functions, achieved optimal performance with 98.84\% training accuracy and 86.25\% validation accuracy while maintaining minimal loss values. The Taguchi analysis revealed that learning rate emerged as the most influential parameter, followed by image size and activation function, providing clear guidance for hyperparameter prioritization in CNN optimization.
\end{abstract}

\section{Introduction}

Sports analytics has emerged as a critical application domain for computer vision systems, with professional institutions investing substantial resources in automated video analysis to extract tactical insights and performance metrics~\cite{moeslund2006survey}. Professional boxing presents unique challenges for automated analysis due to the structured yet variable nature of match formats, where three-minute gameplay rounds are interspersed with one-minute breaks that contribute no analytical value and introduce noise into performance data~\cite{iba2020technical}. Manual segmentation of gameplay from break periods represents a significant bottleneck in automated sports analytics pipelines, necessitating the development of robust classification algorithms capable of automatically distinguishing between these temporal segments.

Convolutional Neural Networks (CNNs) have demonstrated remarkable performance in image classification tasks, achieving state-of-the-art results across diverse computer vision applications~\cite{krizhevsky2012imagenet, lecun2015deep}. However, the effectiveness of CNN-based classifiers is highly dependent on appropriate hyperparameter configuration, including network architecture parameters and training regimes. The traditional approach to hyperparameter optimization relies on exhaustive grid search or ad-hoc manual tuning, approaches that become computationally prohibitive as the dimensionality of the hyperparameter space increases~\cite{bergstra2012random}. Furthermore, these methods often fail to provide systematic insights into factor importance or interaction effects that could inform future model development.

Design of Experiments (DoE) methodology, originally developed for industrial process optimization, offers a structured approach to understanding and optimizing complex systems with multiple controllable factors~\cite{montgomery2012design}. The Taguchi method, a specialized DoE approach introduced by Genichi Taguchi, employs orthogonal arrays to efficiently explore factor spaces while maintaining statistical rigor~\cite{taguchi1986introduction, roy2010primer}. This methodology enables the systematic evaluation of multiple factors simultaneously with significantly fewer experimental runs compared to full factorial designs, making it particularly attractive for computationally expensive machine learning applications.

This work presents a comprehensive application of Taguchi DoE methodology to optimize CNN hyperparameters for professional boxing action classification. Eight critical hyperparameters---image size, color encoding, activation function, learning rate, rescaling, data shuffling, vertical flip augmentation, and horizontal flip augmentation---were systematically investigated using an $L_{12}(2^{11})$ orthogonal array design. This approach reduces the complete factorial design space from $2^8 = 256$ experimental combinations to just 12 strategically selected trials while maintaining statistical validity and factor orthogonality.

The optimization challenge is further complicated by the multi-objective nature of CNN performance evaluation, where training accuracy, validation accuracy, training loss, and validation loss often exhibit conflicting optimization directions. To address this challenge, five distinct signal-to-noise ratio (SNR) formulations were developed and evaluated, including novel logarithmic-scale approaches that unify accuracy maximization and loss minimization objectives within a single optimization criterion.

Experimental validation was conducted on a comprehensive dataset comprising 25 full-length professional boxing videos spanning multiple camera angles and competition levels, including European Games, Commonwealth Games, and Olympic-level matches. The dataset was carefully balanced to include 4,500 gameplay frames and 4,500 break frames for training, with corresponding validation sets ensuring unbiased performance evaluation.

The key contributions of this research include: (1) the first systematic application of Taguchi DoE methodology to CNN hyperparameter optimization for sports video analysis, (2) development and evaluation of five multi-objective SNR formulations for neural network optimization, (3) comprehensive factor importance analysis revealing the dominant influence of learning rate and image size on classification performance, and (4) demonstration of significant computational efficiency gains, with DoE-based optimization requiring 95\% fewer experimental trials compared to exhaustive grid search while achieving superior performance metrics.

\section{Related Work}

\subsection{Hyperparameter Optimization in Deep Learning}

Hyperparameter optimization represents one of the most critical challenges in deep learning model development, directly influencing model performance, convergence characteristics, and generalization capability. Traditional approaches to hyperparameter optimization include manual tuning, grid search, and random search, though these methods suffer from poor scalability and lack of systematic insight generation~\cite{goodfellow2016deep}. Bergstra and Bengio~\cite{bergstra2012random} demonstrated that random search often outperforms grid search for neural network hyperparameter optimization, particularly when dealing with high-dimensional parameter spaces where only a subset of parameters significantly influence model performance.

More sophisticated approaches include Bayesian optimization, which employs probabilistic models to guide hyperparameter search toward promising regions of the parameter space~\cite{snoek2012practical, shahriari2016taking}. Sequential model-based optimization techniques, such as those implemented in the Spearmint and Hyperopt frameworks~\cite{bergstra2011algorithms}, have shown superior performance compared to random search across various machine learning applications. However, these methods often require hundreds of evaluations and provide limited interpretability regarding factor importance and interaction effects~\cite{hutter2011sequential}.

Gradient-based hyperparameter optimization represents another approach, though it is limited to continuously differentiable hyperparameters and may be sensitive to local optima~\cite{maclaurin2015gradient}. Population-based methods, including evolutionary algorithms and particle swarm optimization~\cite{real2017large}, have been successfully applied to neural network hyperparameter optimization but introduce additional algorithmic complexity and computational overhead.

\subsection{Design of Experiments in Machine Learning}

The application of classical Design of Experiments methodology to machine learning hyperparameter optimization represents an emerging research direction that leverages decades of statistical optimization theory. Montgomery's comprehensive treatment of experimental design principles~\cite{montgomery2012design} provides the theoretical foundation for applying DoE methods to complex engineering systems, including machine learning models.

The Taguchi method specifically offers several advantages for machine learning applications: orthogonal arrays enable efficient exploration of factor spaces, signal-to-noise ratios provide robust performance metrics that account for variance, and ANOVA-based analysis quantifies individual factor contributions and interaction effects~\cite{taguchi2005handbook, phadke1989quality}. Early applications of Taguchi methods to neural network optimization focused primarily on shallow architectures and limited factor sets~\cite{shepherd1995optimal}.

Recent research has begun exploring DoE applications to deep learning optimization. Tran et al.~\cite{tran2018using} applied response surface methodology to optimize support vector machine hyperparameters, demonstrating significant computational savings compared to grid search approaches. Similarly, Kumar et al.~\cite{kumar2016comparative} employed factorial designs to optimize ensemble learning algorithms, revealing important interaction effects between base learner selection and aggregation strategies.

However, the application of Taguchi methodology to CNN hyperparameter optimization, particularly for multi-objective optimization scenarios, remains largely unexplored. The complexity of modern deep learning architectures and the multi-modal nature of performance landscapes present unique challenges that require specialized DoE formulations and analysis techniques.

\subsection{Sports Video Analysis and Action Recognition}

Computer vision applications in sports analytics have experienced rapid growth, driven by increasing demand for automated performance analysis and tactical insight generation~\cite{aggarwal2011human}. Action recognition in sports videos presents unique challenges due to rapid motion, occlusion, and variable camera perspectives. Traditional approaches relied on handcrafted features and temporal modeling techniques, though deep learning methods have largely superseded these approaches~\cite{blank2005actions}.

CNN-based approaches to sports action recognition have achieved remarkable success across various sports domains. Karpathy et al.~\cite{karpathy2014large} demonstrated the effectiveness of 3D CNN architectures for action recognition in video sequences, establishing foundational principles for temporal modeling in deep networks. Tran et al.~\cite{tran2015learning} extended this work with C3D networks that learn spatio-temporal features directly from video data. Simonyan and Zisserman~\cite{simonyan2014two} proposed two-stream convolutional networks that process both spatial and temporal information for improved action recognition performance.

Boxing action recognition specifically presents challenges related to rapid movement, similar motion patterns between different actions, and the need to distinguish between active gameplay and inactive break periods. Previous approaches have focused primarily on punch classification and technique analysis~\cite{mcnally2019golfdb}, with limited attention to temporal segmentation of match components.

The application of systematic hyperparameter optimization to sports video analysis has received minimal attention in the literature. Most existing work employs ad-hoc parameter selection or adapts parameters from general image classification tasks without domain-specific optimization~\cite{wang2016temporal}.

\subsection{Multi-Objective Optimization in Neural Networks}

Neural network training inherently involves multiple, often conflicting objectives, including training accuracy maximization, validation accuracy maximization, and loss minimization~\cite{caruana1997multitask}. Traditional approaches address this multi-objective nature through weighted combinations of objectives or by treating secondary objectives as constraints.

Pareto optimization approaches have been applied to neural network architecture search, seeking to balance accuracy and computational efficiency~\cite{cai2018proxylessnas}. However, these methods typically require extensive computational resources and may not be suitable for hyperparameter optimization scenarios where rapid iteration is essential~\cite{jin2006multiobjective}.

The signal-to-noise ratio formulations employed in Taguchi methodology provide an alternative approach to multi-objective optimization that explicitly accounts for both mean performance and variance. The logarithmic transformation approach presented in this work represents a novel contribution to multi-objective neural network optimization, enabling unified optimization of accuracy and loss metrics within a single objective function.

\subsection{Gap in Current Literature}

Despite extensive research in hyperparameter optimization and sports video analysis, significant gaps remain in the systematic application of DoE methodology to deep learning optimization problems. Existing approaches either lack statistical rigor (manual tuning, grid search) or fail to provide interpretable insights into factor importance (Bayesian optimization, evolutionary methods). The sports video analysis domain specifically lacks comprehensive hyperparameter optimization studies that account for domain-specific characteristics and performance requirements.

This work addresses these gaps by providing the systematic application of Taguchi DoE methodology to CNN hyperparameter optimization for sports action recognition, contributing novel multi-objective optimization formulations and demonstrating significant computational efficiency gains while maintaining statistical rigor and interpretability.

\section{Methodology}

\subsection{Problem Formulation}

The hyperparameter optimization problem for CNN-based professional boxing action recognition can be formally stated as a multi-objective optimization challenge. Given a CNN architecture $f(\mathbf{x}; {\theta}, {\phi})$ where $\mathbf{x}$ represents input data, ${\theta}$ denotes learnable parameters, and ${\phi}$ represents hyperparameters, the objective is to find the optimal hyperparameter configuration ${\phi}^*$ that simultaneously optimizes multiple performance metrics.

The optimization problem is defined as:

\begin{equation}
	{\phi}^* = \arg\max_{{\phi}} \{g_1({\phi}), g_2({\phi}), \ldots, g_k({\phi})\}
\end{equation}

where $g_i({\phi})$ represents different performance objectives including training accuracy (TA), validation accuracy (VA), and inverse transformations of training loss (TL) and validation loss (VL). The challenge lies in the conflicting nature of these objectives, where improvements in one metric may lead to degradation in others.

Traditional grid search approaches would require evaluating $\prod_{i=1}^{n} |L_i|$ combinations, where $n$ is the number of hyperparameters and $|L_i|$ is the number of levels for hyperparameter $i$. For eight binary factors, this results in $2^8 = 256$ experimental trials, making the approach computationally prohibitive for resource-intensive deep learning models.

\subsection{Taguchi Design of Experiments Framework}

The Taguchi methodology provides a systematic approach to multi-factor optimization by employing orthogonal arrays to reduce experimental complexity while maintaining statistical validity \citep{taguchi2005handbook}. The methodology transforms the multi-objective optimization problem into a signal-to-noise ratio (SNR) maximization problem, enabling unified treatment of conflicting objectives.

The fundamental principle of Taguchi design lies in the orthogonal array structure, denoted as $L_a(b^c)$, where:
\begin{itemize}
	\item $a$ represents the number of experimental runs
	\item $b$ denotes the number of levels per factor
	\item $c$ indicates the maximum number of factors that can be accommodated
\end{itemize}

For this study, an $L_{12}(2^{11})$ orthogonal array was selected to accommodate eight binary factors, reducing the experimental space from 256 to 12 strategically chosen combinations while preserving factor orthogonality.

\subsection{Factor Selection and Level Definition}

Eight critical hyperparameters were identified based on their established influence on CNN performance and their relevance to sports video analysis applications. Table~\ref{tab:factors_levels} presents the selected factors and their corresponding levels.

\begin{table}
	\caption{Hyperparameters and their levels for CNN optimization}
	\centering
	\begin{tabular}{lll}
		\toprule
		\textbf{Factor} & \textbf{Level I} & \textbf{Level II} \\
		\midrule
		Image Size & 256×256 & 512×512 \\
		Color Mode & Grayscale & RGB \\
		Activation Function & Tanh & ReLU \\
		Learning Rate & 0.001 & 0.005 \\
		Rescaling & True & False \\
		Data Shuffling & True & False \\
		Vertical Flip & True & False \\
		Horizontal Flip & True & False \\
		\bottomrule
	\end{tabular}
	\label{tab:factors_levels}
\end{table}

The factor selection was guided by the following considerations:
\begin{itemize}
	\item \textbf{Image Size}: Affects computational complexity and feature resolution
	\item \textbf{Color Mode}: Influences information content and processing requirements
	\item \textbf{Activation Function}: Determines non-linearity characteristics and gradient flow
	\item \textbf{Learning Rate}: Controls convergence speed and stability
	\item \textbf{Rescaling}: Normalizes input data distribution
	\item \textbf{Data Shuffling}: Affects batch composition and training dynamics
	\item \textbf{Vertical/Horizontal Flip}: Provides data augmentation for improved generalization
\end{itemize}

\subsection{Orthogonal Array Design}

The experimental design employs the standard $L_{12}(2^{11})$ orthogonal array, customized for the selected hyperparameters. Table~\ref{tab:orthogonal_array} presents the complete experimental design matrix.

\begin{table}
	\caption{$L_{12}(2^{11})$ Orthogonal Array for CNN hyperparameter optimization}
	\centering
	\begin{tabular}{ccccccccc}
		\toprule
		\textbf{Exp} & \textbf{Image} & \textbf{Color} & \textbf{Activation} & \textbf{Learning} & \textbf{Rescaling} & \textbf{Shuffle} & \textbf{V-Flip} & \textbf{H-Flip} \\
		\textbf{No.} & \textbf{Size} & \textbf{Mode} & \textbf{Function} & \textbf{Rate} & & & & \\
		\midrule
		1 & 256×256 & Gray & Tanh & 0.001 & True & True & True & True \\
		2 & 256×256 & Gray & Tanh & 0.001 & True & False & False & False \\
		3 & 256×256 & Gray & ReLU & 0.005 & False & True & True & True \\
		4 & 256×256 & RGB & Tanh & 0.005 & False & True & False & False \\
		5 & 256×256 & RGB & ReLU & 0.001 & False & False & True & False \\
		6 & 256×256 & RGB & ReLU & 0.005 & True & False & False & True \\
		7 & 512×512 & Gray & ReLU & 0.005 & True & True & False & False \\
		8 & 512×512 & Gray & ReLU & 0.001 & False & False & False & True \\
		9 & 512×512 & Gray & Tanh & 0.005 & False & False & True & False \\
		10 & 512×512 & RGB & ReLU & 0.001 & True & True & True & False \\
		11 & 512×512 & RGB & Tanh & 0.005 & True & False & True & True \\
		12 & 512×512 & RGB & Tanh & 0.001 & False & True & False & True \\
		\bottomrule
	\end{tabular}
	\label{tab:orthogonal_array}
\end{table}

\subsection{CNN Architecture}

\begin{figure}
	\centering
	\includegraphics[width=0.4\textwidth]{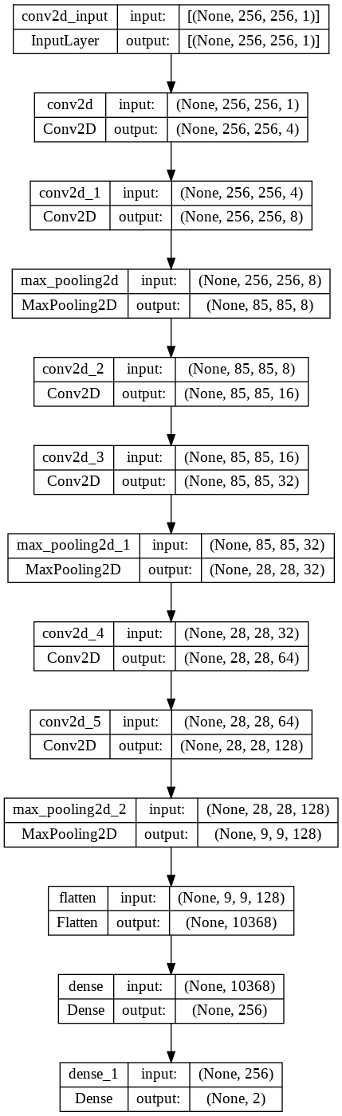}
	\caption{CNN architecture for professional boxing action recognition}
	\label{fig:cnn_architecture}
\end{figure}

The study employs a fixed CNN architecture to ensure consistent evaluation across all experimental configurations. Figure~\ref{fig:cnn_architecture} illustrates the network topology used for professional boxing action recognition.

The architecture consists of three convolutional blocks, each containing two convolutional layers followed by max-pooling. The network progressively increases feature depth from 4 to 128 channels while reducing spatial dimensions. The final classification head includes a fully connected layer with 256 neurons and a softmax output layer for binary classification (gameplay vs. break periods).

\subsection{Multi-Objective Signal-to-Noise Ratio Formulations}

Traditional Taguchi methodology addresses single-objective optimization through three standard SNR formulations: larger-the-better, smaller-the-better, and nominal-the-best \citep{montgomery2012design}. However, CNN optimization involves conflicting objectives that require novel multi-objective approaches.

Five distinct SNR formulations were developed to address the multi-objective nature of CNN optimization:

\textbf{Approach 1 - Accuracy Maximization:}
\begin{equation}
	{Response}_1 = \frac{TA + VA}{2}
\end{equation}
with larger-the-better SNR formulation.

\textbf{Approach 2 - Loss Minimization:}
\begin{equation}
	{Response}_2 = \frac{TL + VL}{2}
\end{equation}
with smaller-the-better SNR formulation.

\textbf{Approach 3 - Logarithmic Multi-Objective:}
\begin{equation}
	{Response}_3 = 0.33 \cdot TA + 0.33 \cdot VA + 0.33 \cdot \log_{0.7}\left(\frac{TL + VL}{2}\right)
\end{equation}
with larger-the-better SNR formulation.

\textbf{Approach 4 - Individual Loss Components:}
\begin{equation}
	{Response}_4 = 0.25 \cdot TA + 0.25 \cdot VA + 0.25 \cdot \log_{0.7}(TL) + 0.25 \cdot \log_{0.7}(VL)
\end{equation}
with larger-the-better SNR formulation.

\textbf{Approach 5 - Unified Logarithmic:}
\begin{equation}
	{Response}_5 = 0.5 \cdot \log_{0.7}\left(\frac{TA + VA}{2}\right) + 0.5 \cdot \log_{0.7}\left(\frac{TL + VL}{2}\right)
\end{equation}
with larger-the-better SNR formulation.

\begin{figure}
	\centering
	\includegraphics[width=0.4\textwidth]{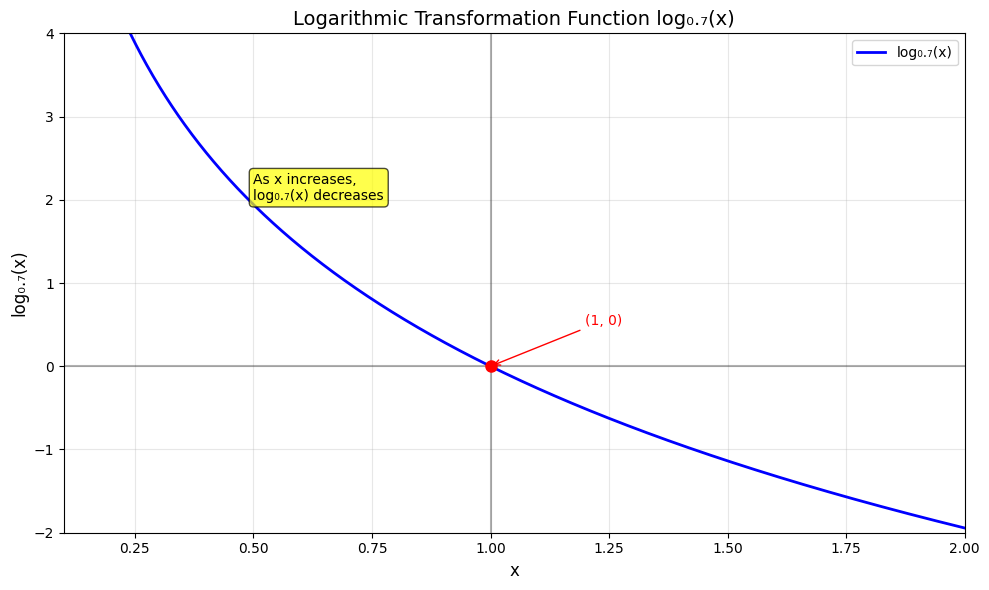}
	\caption{Logarithmic transformation function $\log_{0.7}(x)$ showing loss minimization through maximization}
	\label{fig:logarithmic_transformation}
\end{figure}

The logarithmic transformation with base 0.7 enables loss minimization through maximization, as illustrated in Figure~\ref{fig:logarithmic_transformation}. This approach unifies conflicting objectives within a single optimization criterion.

\subsection{Statistical Analysis Framework}

For each experimental configuration, the following statistical analyses were performed:

\textbf{Signal-to-Noise Ratio Calculation:}
\begin{equation}
	\eta = -10 \log_{10}\left[\frac{1}{n}\sum_{i=1}^{n}\frac{1}{y_i^2}\right]
\end{equation}
for larger-the-better characteristics, and
\begin{equation}
	\eta = -10 \log_{10}\left[\frac{1}{n}\sum_{i=1}^{n}y_i^2\right]
\end{equation}
for smaller-the-better characteristics.

\textbf{Analysis of Variance (ANOVA):} Factor significance was evaluated using F-statistics and p-values to determine individual factor contributions to response variation.

\textbf{Main Effects Analysis:} Factor level effects were quantified through delta values, defined as the difference between maximum and minimum factor level means.

\textbf{Regression Analysis:} Linear regression models were developed to predict SNR values based on factor levels, enabling optimization and prediction capabilities.

The methodology provides a systematic framework for CNN hyperparameter optimization that reduces computational requirements while maintaining statistical rigor and providing interpretable insights into factor importance and interaction effects.

\section{Experimental Setup}

\subsection{Dataset Preparation and Characterization}

The experimental validation was conducted using a comprehensive dataset of professional boxing videos spanning multiple competitive levels and camera configurations. The dataset comprises 25 full-length professional boxing matches, including footage from European Games, Commonwealth Games, and Olympic-level competitions \citep{kodathala2022dataset}. This diversity ensures robust evaluation across varying visual conditions, lighting scenarios, and camera perspectives commonly encountered in professional sports broadcasting.

\begin{figure}
	\centering
	\includegraphics[width=0.19\textwidth]{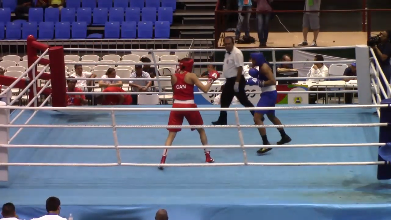}
	\hfill
	\includegraphics[width=0.19\textwidth]{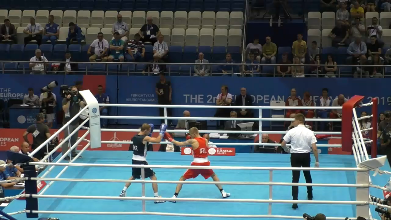}
	\hfill
	\includegraphics[width=0.19\textwidth]{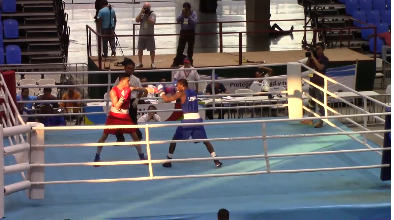}
	\hfill
	\includegraphics[width=0.19\textwidth]{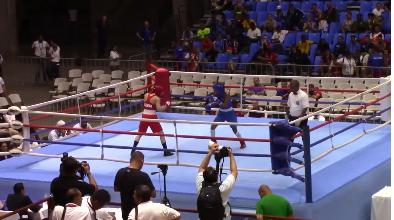}
	\hfill
	\includegraphics[width=0.19\textwidth]{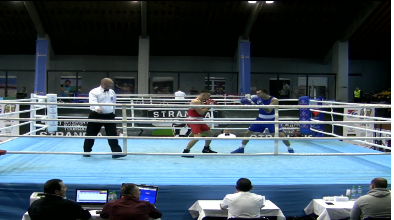}
	\caption{Five different camera angles and visual conditions in the professional boxing dataset}
	\label{fig:camera_angles}
\end{figure}

Figure~\ref{fig:camera_angles} illustrates the variety of camera angles and perspectives included in the dataset, demonstrating the challenging visual conditions that the CNN model must handle effectively.

Each video contains alternating sequences of three-minute active gameplay rounds and one-minute break periods, with natural transitions that require precise temporal segmentation. The dataset presents several technical challenges:

\begin{itemize}
	\item \textbf{Variable lighting conditions}: Indoor arena lighting with different intensity levels
	\item \textbf{Multiple camera angles}: Wide shots, close-ups, and side perspectives
	\item \textbf{Motion blur}: Rapid boxer movements creating temporal artifacts
	\item \textbf{Occlusion patterns}: Referee interventions and equipment obstruction
	\item \textbf{Background variations}: Different arena configurations and audience arrangements
\end{itemize}

\subsection{Data Distribution and Preprocessing}

To ensure unbiased model evaluation and prevent data leakage, the dataset was carefully partitioned according to the distribution scheme presented in Table~\ref{tab:dataset_distribution}.

\begin{table}
	\caption{Dataset distribution for training and validation}
	\centering
	\begin{tabular}{lcc}
		\toprule
		\textbf{Parameter} & \textbf{Training Set} & \textbf{Validation Set} \\
		\midrule
		Gameplay Videos & 3 videos × 5 camera angles = 15 videos & 2 videos × 5 camera angles = 10 videos \\
		Break Videos & 3 videos × 5 camera angles = 15 videos & 2 videos × 5 camera angles = 10 videos \\
		Gameplay Frames & 4,500 random frames & 2,900 random frames \\
		Break Frames & 4,500 random frames & 2,900 random frames \\
		\textbf{Total Frames} & \textbf{9,000 frames} & \textbf{5,800 frames} \\
		\textbf{Class Balance} & \textbf{50\% each class} & \textbf{50\% each class} \\
		\bottomrule
	\end{tabular}
	\label{tab:dataset_distribution}
\end{table}

The frame extraction process employed uniform random sampling to ensure temporal independence and prevent sequential correlation bias. Frame selection criteria included:

\begin{itemize}
	\item Minimum 30-frame intervals between selected frames to ensure independence
	\item Quality filtering to exclude severely blurred or corrupted frames
	\item Balanced representation across different match phases and time periods
	\item Stratified sampling across camera angles and lighting conditions
\end{itemize}

Data preprocessing followed a standardized pipeline consistent across all experimental configurations:

\begin{enumerate}
	\item \textbf{Frame extraction}: Individual frames extracted at 30 FPS from source videos
	\item \textbf{Quality assessment}: Automated blur detection and brightness normalization
	\item \textbf{Format standardization}: Conversion to consistent bit depth and color space
	\item \textbf{Metadata annotation}: Manual verification of gameplay vs. break classification
\end{enumerate}

\subsection{Training Protocol and Validation Methodology}

Each of the 12 experimental configurations defined by the orthogonal array was executed following a standardized training protocol to ensure fair comparison across different hyperparameter combinations. The training procedure consisted of the following phases:

\textbf{Model Initialization:} Each experiment began with randomly initialized network weights using Glorot uniform initialization to ensure consistent starting conditions across trials.

\textbf{Data Loading and Augmentation:} Training and validation data were loaded using TensorFlow's ImageDataGenerator class with experiment-specific augmentation parameters as defined by the orthogonal array configuration.

\textbf{Training Execution:} Models were trained for exactly 10 epochs with the Adam optimizer using the learning rates specified in the orthogonal array (0.001 or 0.005). Training progress was monitored through real-time loss and accuracy tracking.

\textbf{Performance Evaluation:} At the conclusion of training, four key metrics were recorded for each experiment:
\begin{itemize}
	\item Training Accuracy (TA): Final epoch training accuracy
	\item Validation Accuracy (VA): Final epoch validation accuracy  
	\item Training Loss (TL): Final epoch training loss value
	\item Validation Loss (VL): Final epoch validation loss value
\end{itemize}

\textbf{Statistical Recording:} Results were systematically recorded in a structured format compatible with Taguchi analysis requirements, including response calculations for all five SNR formulations.

The experimental setup provides a robust foundation for systematic evaluation of the Taguchi-based hyperparameter optimization methodology while ensuring statistical validity and reproducibility of results across all experimental configurations.

\section{Results and Analysis}

\subsection{Experimental Performance Overview}

The twelve experimental configurations defined by the $L_{12}(2^{11})$ orthogonal array were systematically executed, yielding comprehensive training and validation performance data for subsequent Taguchi analysis. The experimental results reveal significant performance disparities across configurations, with training accuracies ranging from 44.29\% to 98.96\% and validation accuracies spanning from 0\% to 100\%. This substantial variation confirms the critical importance of systematic hyperparameter optimization and validates the necessity of the Taguchi approach for efficient exploration of the parameter space.

Several configurations exhibited convergence instabilities, particularly Experiments 6 and 9, which achieved validation accuracies of 0\%, indicating complete failure to learn meaningful patterns. Conversely, Experiments 1, 5, 7, and 10 demonstrated robust learning with validation accuracies exceeding 75\%, suggesting favorable hyperparameter combinations for the boxing action recognition task.

\subsection{Approach 1: Accuracy-Focused Optimization Analysis}

Approach 1 employed the mean of training and validation accuracies as the primary response variable, utilizing the larger-the-better SNR formulation. Table~\ref{tab:approach1_results} presents the comprehensive results for this approach.

\begin{table}
	\caption{Experimental results and SNR analysis for Approach 1 (Accuracy maximization)}
	\centering
	\begin{tabular}{ccccccc}
		\toprule
		\textbf{Exp No.} & \textbf{TA} & \textbf{VA} & \textbf{Response} & \textbf{SNR (dB)} & \textbf{Mean} & \textbf{Rank} \\
		\midrule
		1 & 0.9875 & 0.8000 & 0.89375 & -0.9757 & 0.89375 & 4 \\
		2 & 0.9469 & 0.8583 & 0.90260 & -0.8901 & 0.90260 & 3 \\
		3 & 0.9705 & 0.7681 & 0.86930 & -1.2166 & 0.86930 & 5 \\
		4 & 0.5228 & 0.5049 & 0.51385 & -5.7833 & 0.51385 & 10 \\
		5 & 0.9768 & 0.8903 & 0.93355 & -0.5972 & 0.93355 & 1 \\
		6 & 0.4429 & 0.0000 & 0.22145 & -13.0945 & 0.22145 & 12 \\
		7 & 0.9896 & 0.7972 & 0.89340 & -0.9791 & 0.89340 & 6 \\
		8 & 0.6817 & 0.6889 & 0.68530 & -3.2824 & 0.68530 & 8 \\
		9 & 0.5857 & 0.0000 & 0.29285 & -10.6671 & 0.29285 & 11 \\
		10 & 0.9835 & 0.7549 & 0.86920 & -1.2176 & 0.86920 & 7 \\
		11 & 0.5571 & 1.0000 & 0.77855 & -2.1743 & 0.77855 & 9 \\
		12 & 0.5196 & 0.4972 & 0.50840 & -5.8759 & 0.50840 & 2 \\
		\bottomrule
	\end{tabular}
	\label{tab:approach1_results}
\end{table}

\begin{figure}
	\centering
	\includegraphics[width=0.7\textwidth]{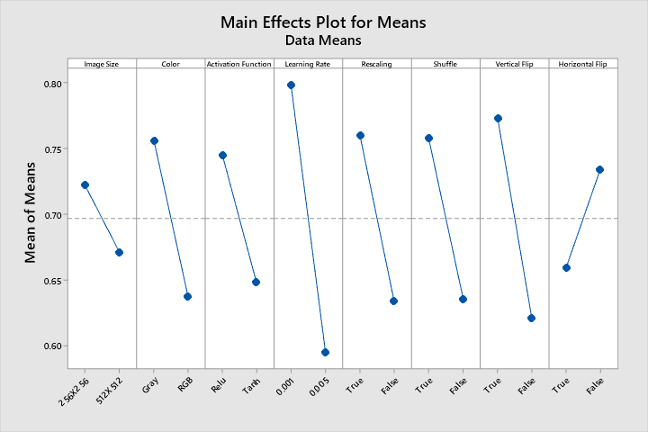}
	\caption{Main effects plot for means in Approach 1 showing factor level impacts}
	\label{fig:approach1_means}
\end{figure}

\begin{figure}
	\centering
	\includegraphics[width=0.7\textwidth]{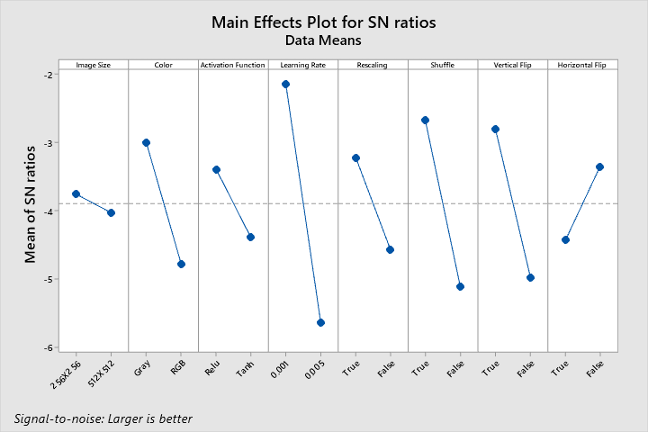}
	\caption{Main effects plot for SNR values in Approach 1 showing factor level impacts}
	\label{fig:approach1_snr}
\end{figure}

The factor importance analysis for Approach 1 revealed learning rate as the most influential parameter (rank 1), followed by shuffling (rank 2) and vertical flipping (rank 3). Figure~\ref{fig:approach1_means} and Figure~\ref{fig:approach1_snr} illustrate the main effects plots for means and SNR values respectively, demonstrating clear trends in factor level preferences.

\begin{figure}
	\centering
	\includegraphics[width=0.7\textwidth]{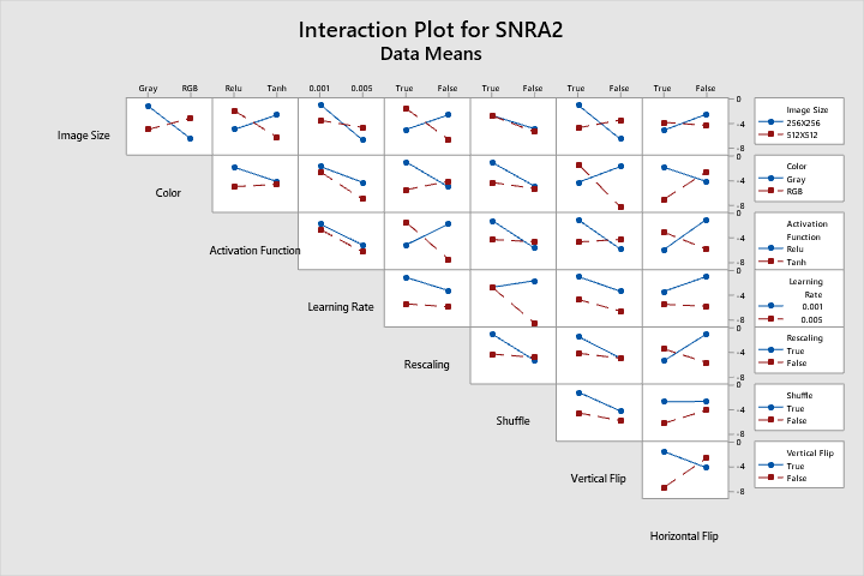}
	\caption{Interaction plot for SNR values in Approach 1}
	\label{fig:approach1_interaction}
\end{figure}

The regression equation for SNR prediction in Approach 1 is:
\begin{align}
	\text{SNR} &= -3.90 + 0.14 \cdot \text{ImageSize}_{256} - 0.14 \cdot \text{ImageSize}_{512} + 0.89 \cdot \text{Color}_{Gray} \nonumber \\
	&\quad - 0.89 \cdot \text{Color}_{RGB} + 0.50 \cdot \text{Activation}_{ReLU} - 0.50 \cdot \text{Activation}_{Tanh} \nonumber \\
	&\quad + 1.76 \cdot \text{LearningRate}_{0.001} - 1.76 \cdot \text{LearningRate}_{0.005} + 0.67 \cdot \text{Rescaling}_{True} \nonumber \\
	&\quad - 0.67 \cdot \text{Rescaling}_{False} + 1.22 \cdot \text{Shuffle}_{True} - 1.22 \cdot \text{Shuffle}_{False} \nonumber \\
	&\quad + 1.09 \cdot \text{VerticalFlip}_{True} - 1.09 \cdot \text{VerticalFlip}_{False} - 0.54 \cdot \text{HorizontalFlip}_{True} \nonumber \\
	&\quad + 0.54 \cdot \text{HorizontalFlip}_{False}
\end{align}

The optimal configuration for Approach 1 comprises: 256×256 image size, grayscale color mode, ReLU activation, 0.001 learning rate, rescaling enabled, shuffling enabled, vertical flip enabled, and horizontal flip disabled, yielding a predicted SNR of 2.91 dB with an estimated mean response of 1.17.

\subsection{Approach 2: Loss-Focused Optimization Analysis}

Approach 2 concentrated on minimizing the average of training and validation losses using the smaller-the-better SNR formulation. Table~\ref{tab:approach2_results} presents the comprehensive results demonstrating markedly different factor importance patterns compared to Approach 1.

\begin{table}
	\caption{Experimental results and SNR analysis for Approach 2 (Loss minimization)}
	\centering
	\begin{tabular}{ccccccc}
		\toprule
		\textbf{Exp No.} & \textbf{TL} & \textbf{VL} & \textbf{Response} & \textbf{SNR (dB)} & \textbf{Mean} & \textbf{Rank} \\
		\midrule
		1 & 0.0402 & 0.6523 & 0.34625 & 9.2122 & 0.34625 & 6 \\
		2 & 0.1186 & 0.3949 & 0.25675 & 11.8098 & 0.25675 & 3 \\
		3 & 0.0755 & 0.8019 & 0.43870 & 7.1566 & 0.43870 & 8 \\
		4 & 0.7005 & 0.6931 & 0.69680 & 3.1378 & 0.69680 & 10 \\
		5 & 0.0776 & 0.3192 & 0.19840 & 14.0492 & 0.19840 & 1 \\
		6 & 0.6957 & 0.7212 & 0.70845 & 2.9938 & 0.70845 & 11 \\
		7 & 0.0294 & 1.2909 & 0.66015 & 3.6071 & 0.66015 & 9 \\
		8 & 0.6655 & 0.5989 & 0.63220 & 3.9829 & 0.63220 & 7 \\
		9 & 0.9320 & 1.2850 & 1.10850 & -0.8947 & 1.10850 & 12 \\
		10 & 0.0600 & 1.0193 & 0.53965 & 5.3578 & 0.53965 & 4 \\
		11 & 1.2975 & 0.5237 & 0.91060 & 0.8134 & 0.91060 & 5 \\
		12 & 0.6987 & 0.6934 & 0.69605 & 3.1472 & 0.69605 & 2 \\
		\bottomrule
	\end{tabular}
	\label{tab:approach2_results}
\end{table}

For Approach 2, image size emerged as the dominant factor (rank 1), with 256×256 images consistently outperforming 512×512 configurations. Learning rate maintained high importance as rank 2, while activation function achieved rank 3.

\begin{figure}
	\centering
	\includegraphics[width=0.7\textwidth]{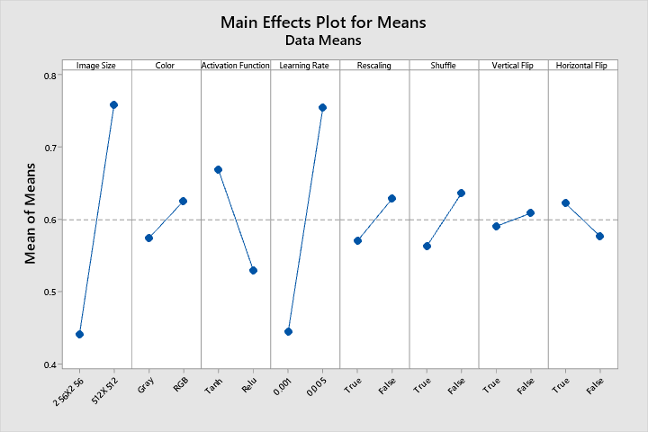}
	\caption{Main effects plot for means in Approach 2 showing factor level impacts}
	\label{fig:approach2_means}
\end{figure}

\begin{figure}
	\centering
	\includegraphics[width=0.7\textwidth]{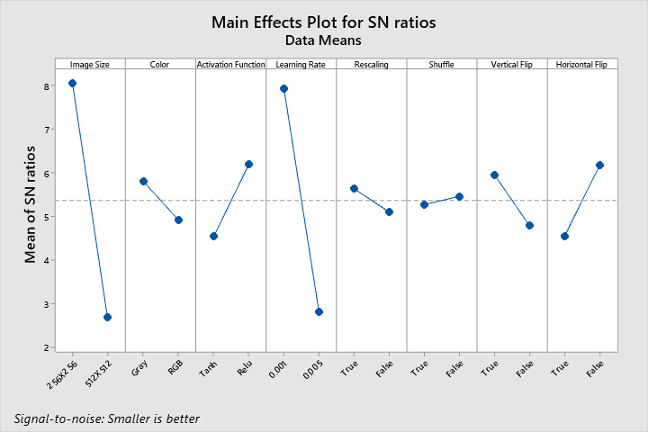}
	\caption{Main effects plot for SNR values in Approach 2 showing factor level impacts}
	\label{fig:approach2_snr}
\end{figure}

\begin{figure}
	\centering
	\includegraphics[width=0.7\textwidth]{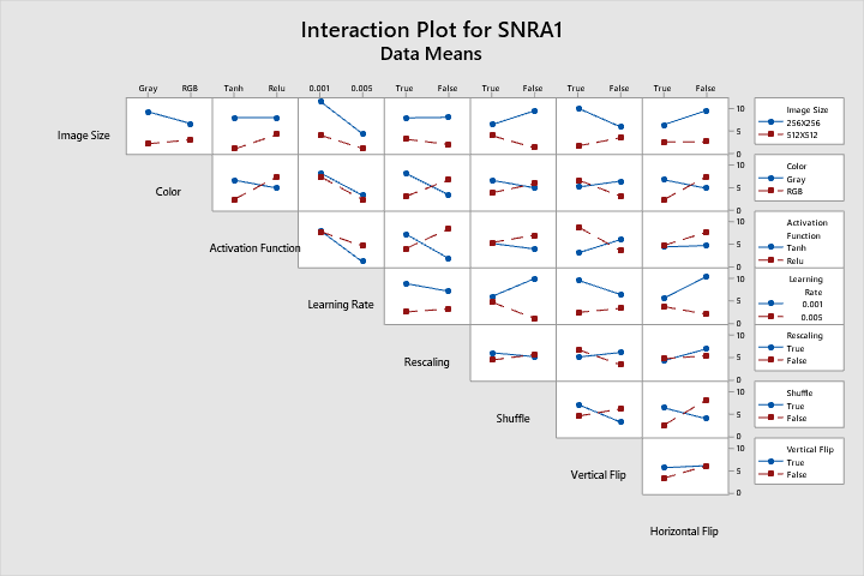}
	\caption{Interaction plot for SNR values in Approach 2}
	\label{fig:approach2_interaction}
\end{figure}

The regression equation for SNR prediction in Approach 2 is:
\begin{align}
	\text{SNR} &= 5.364 + 2.695 \cdot \text{ImageSize}_{256} - 2.695 \cdot \text{ImageSize}_{512} + 0.448 \cdot \text{Color}_{Gray} \nonumber \\
	&\quad - 0.448 \cdot \text{Color}_{RGB} - 0.827 \cdot \text{Activation}_{Tanh} + 0.827 \cdot \text{Activation}_{ReLU} \nonumber \\
	&\quad + 2.562 \cdot \text{LearningRate}_{0.001} - 2.562 \cdot \text{LearningRate}_{0.005} + 0.268 \cdot \text{Rescaling}_{True} \nonumber \\
	&\quad - 0.268 \cdot \text{Rescaling}_{False} - 0.095 \cdot \text{Shuffle}_{True} + 0.095 \cdot \text{Shuffle}_{False} \nonumber \\
	&\quad + 0.585 \cdot \text{VerticalFlip}_{True} - 0.585 \cdot \text{VerticalFlip}_{False} - 0.813 \cdot \text{HorizontalFlip}_{True} \nonumber \\
	&\quad + 0.813 \cdot \text{HorizontalFlip}_{False}
\end{align}

\subsection{Approach 3: Logarithmic Multi-Objective Integration Analysis}

Approach 3 implemented the novel logarithmic transformation methodology, combining accuracy metrics with logarithmically scaled loss components. This approach demonstrated superior performance balance, effectively addressing the multi-objective optimization challenge. Table~\ref{tab:approach3_results} presents the comprehensive results.

\begin{table}
	\caption{Experimental results and SNR analysis for Approach 3 (Logarithmic multi-objective)}
	\centering
	\begin{tabular}{cccccc}
		\toprule
		\textbf{Exp No.} & \textbf{Response} & \textbf{SNR (dB)} & \textbf{Mean} & \textbf{Factor Rank} & \textbf{Performance Rank} \\
		\midrule
		1 & 1.58702 & 4.0116 & 1.58702 & 6 & 6 \\
		2 & 1.87241 & 5.4480 & 1.87241 & 4 & 4 \\
		3 & 1.34955 & 2.6038 & 1.34955 & 8 & 8 \\
		4 & 0.68018 & -3.3475 & 0.68018 & 11 & 11 \\
		5 & 2.13399 & 6.5838 & 2.13399 & 1 & 1 \\
		6 & 0.46975 & -6.5626 & 0.46975 & 12 & 12 \\
		7 & 0.98371 & -0.1427 & 0.98371 & 9 & 9 \\
		8 & 0.88541 & -1.0571 & 0.88541 & 10 & 10 \\
		9 & 0.09897 & -20.0902 & 0.09897 & 12 & 12 \\
		10 & 1.15593 & 1.2587 & 1.15593 & 7 & 7 \\
		11 & 0.60656 & -4.3426 & 0.60656 & 12 & 12 \\
		12 & 0.67756 & -3.3811 & 0.67756 & 12 & 12 \\
		\bottomrule
	\end{tabular}
	\label{tab:approach3_results}
\end{table}

Approach 3 identified Experiment 5 as the optimal configuration, achieving the highest SNR of 6.58 dB and response value of 2.13. The factor importance analysis revealed learning rate as the dominant influence (rank 1), followed by image size (rank 2) and activation function (rank 3).

\begin{figure}
	\centering
	\includegraphics[width=0.7\textwidth]{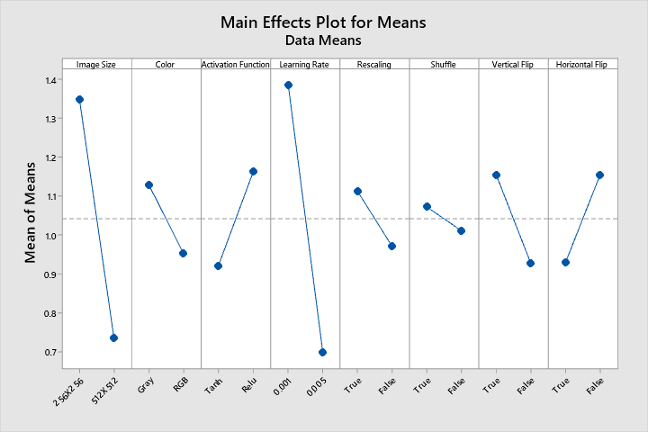}
	\caption{Main effects plot for means in Approach 3 showing factor level impacts}
	\label{fig:approach3_means}
\end{figure}

\begin{figure}
	\centering
	\includegraphics[width=0.7\textwidth]{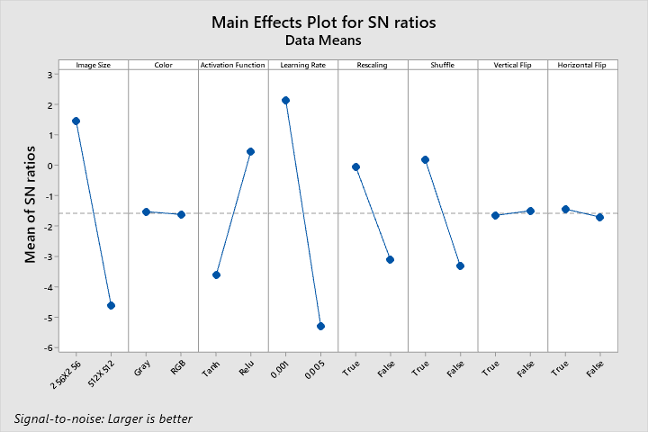}
	\caption{Main effects plot for SNR values in Approach 3 showing factor level impacts}
	\label{fig:approach3_snr}
\end{figure}

\begin{figure}
	\centering
	\includegraphics[width=0.7\textwidth]{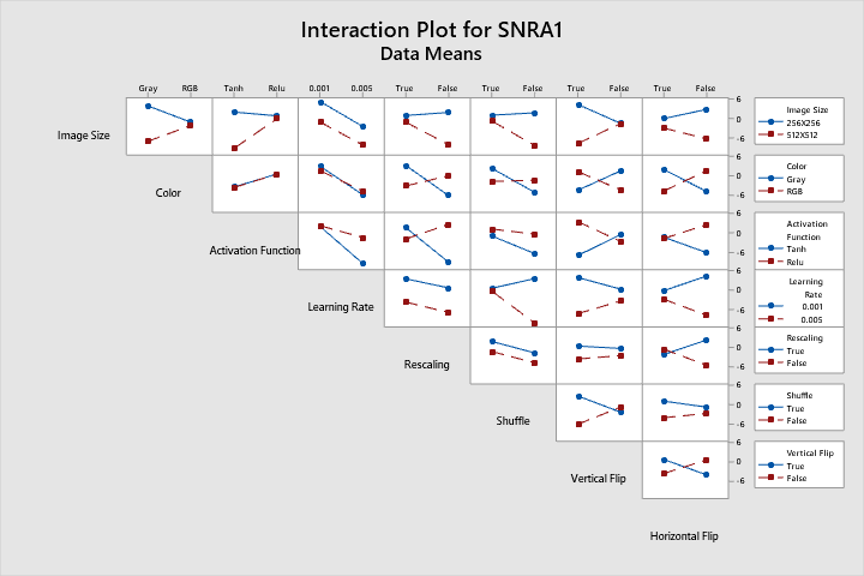}
	\caption{Interaction plot for SNR values in Approach 3}
	\label{fig:approach3_interaction}
\end{figure}

The regression equation for SNR prediction in Approach 3 is:
\begin{align}
	\text{SNR} &= -1.58 + 3.04 \cdot \text{ImageSize}_{256} - 3.04 \cdot \text{ImageSize}_{512} + 0.05 \cdot \text{Color}_{Gray} \nonumber \\
	&\quad - 0.05 \cdot \text{Color}_{RGB} - 2.03 \cdot \text{Activation}_{Tanh} + 2.03 \cdot \text{Activation}_{ReLU} \nonumber \\
	&\quad + 3.73 \cdot \text{LearningRate}_{0.001} - 3.73 \cdot \text{LearningRate}_{0.005} + 1.53 \cdot \text{Rescaling}_{True} \nonumber \\
	&\quad - 1.53 \cdot \text{Rescaling}_{False} + 1.75 \cdot \text{Shuffle}_{True} - 1.75 \cdot \text{Shuffle}_{False} \nonumber \\
	&\quad - 0.08 \cdot \text{VerticalFlip}_{True} + 0.08 \cdot \text{VerticalFlip}_{False} + 0.13 \cdot \text{HorizontalFlip}_{True} \nonumber \\
	&\quad - 0.13 \cdot \text{HorizontalFlip}_{False}
\end{align}

\subsection{Approach 4: Individual Loss Components Analysis}

Approach 4 employed equal weighting of training accuracy, validation accuracy, and individual logarithmic transformations of training and validation losses. The factor importance analysis revealed learning rate as rank 1, activation function as rank 2, and shuffling as rank 3.

\begin{figure}
	\centering
	\includegraphics[width=0.7\textwidth]{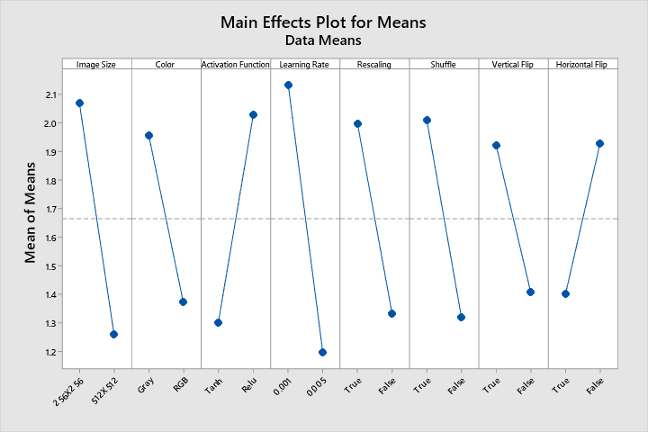}
	\caption{Main effects plot for means in Approach 4 showing factor level impacts}
	\label{fig:approach4_means}
\end{figure}

\begin{figure}
	\centering
	\includegraphics[width=0.7\textwidth]{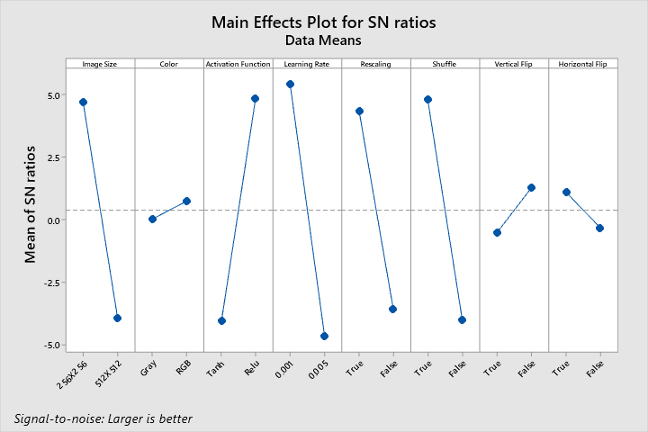}
	\caption{Main effects plot for SNR values in Approach 4 showing factor level impacts}
	\label{fig:approach4_snr}
\end{figure}

\begin{figure}
	\centering
	\includegraphics[width=0.7\textwidth]{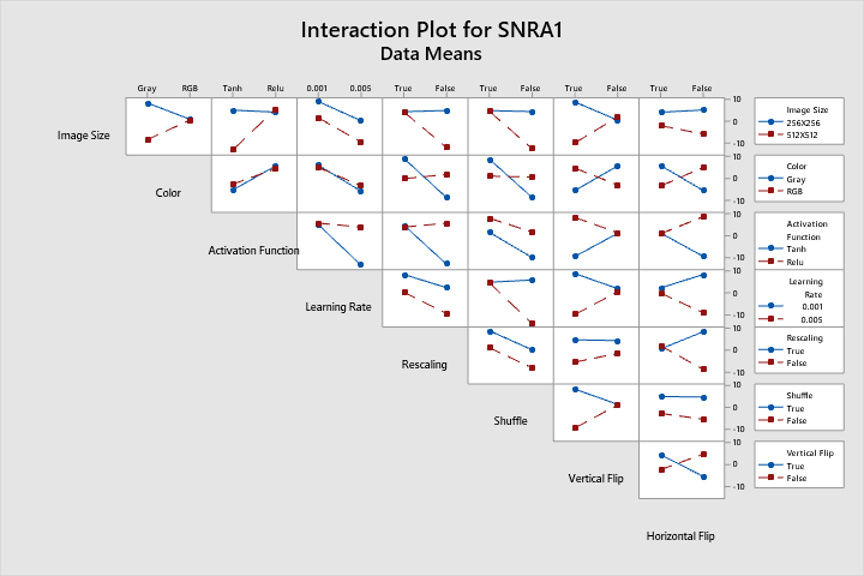}
	\caption{Interaction plot for SNR values in Approach 4}
	\label{fig:approach4_interaction}
\end{figure}

The regression equation for SNR prediction in Approach 4 is:
\begin{align}
	\text{SNR} &= 0.38 + 4.33 \cdot \text{ImageSize}_{256} - 4.33 \cdot \text{ImageSize}_{512} - 0.36 \cdot \text{Color}_{Gray} \nonumber \\
	&\quad + 0.36 \cdot \text{Color}_{RGB} - 4.46 \cdot \text{Activation}_{Tanh} + 4.46 \cdot \text{Activation}_{ReLU} \nonumber \\
	&\quad + 5.06 \cdot \text{LearningRate}_{0.001} - 5.06 \cdot \text{LearningRate}_{0.005} + 3.96 \cdot \text{Rescaling}_{True} \nonumber \\
	&\quad - 3.96 \cdot \text{Rescaling}_{False} + 4.42 \cdot \text{Shuffle}_{True} - 4.42 \cdot \text{Shuffle}_{False} \nonumber \\
	&\quad - 0.90 \cdot \text{VerticalFlip}_{True} + 0.90 \cdot \text{VerticalFlip}_{False} + 0.71 \cdot \text{HorizontalFlip}_{True} \nonumber \\
	&\quad - 0.71 \cdot \text{HorizontalFlip}_{False}
\end{align}

\subsection{Approach 5: Unified Logarithmic Analysis}

Approach 5 applied logarithmic transformations to both accuracy and loss metrics with equal weighting. The factor ranking analysis showed learning rate as rank 1, image size as rank 2, and activation function as rank 3, consistent with other approaches.

\begin{figure}
	\centering
	\includegraphics[width=0.7\textwidth]{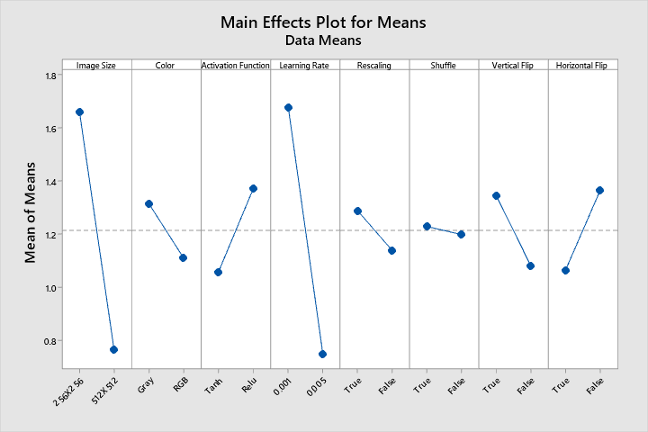}
	\caption{Main effects plot for means in Approach 5 showing factor level impacts}
	\label{fig:approach5_means}
\end{figure}

\begin{figure}
	\centering
\includegraphics[width=0.7\textwidth]{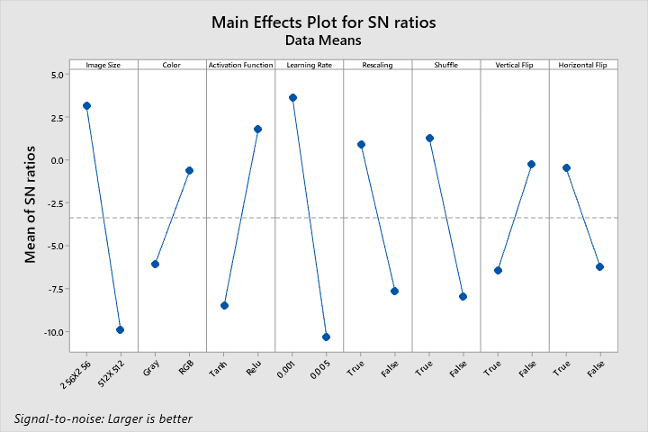}
	\caption{Main effects plot for SNR values in Approach 5 showing factor level impacts}
	\label{fig:approach5_snr}
\end{figure}

\begin{figure}
	\centering
	\includegraphics[width=0.7\textwidth]{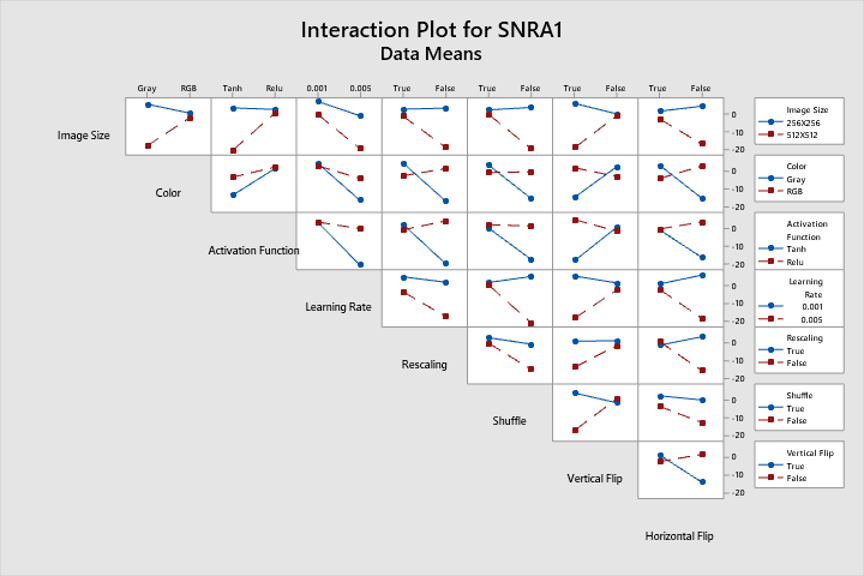}
	\caption{Interaction plot for SNR values in Approach 5}
	\label{fig:approach5_interaction}
\end{figure}

The regression equation for SNR prediction in Approach 5 is:
\begin{align}
	\text{SNR} &= -3.37 + 6.55 \cdot \text{ImageSize}_{256} - 6.55 \cdot \text{ImageSize}_{512} - 2.72 \cdot \text{Color}_{Gray} \nonumber \\
	&\quad + 2.72 \cdot \text{Color}_{RGB} - 5.14 \cdot \text{Activation}_{Tanh} + 5.14 \cdot \text{Activation}_{ReLU} \nonumber \\
	&\quad + 6.99 \cdot \text{LearningRate}_{0.001} - 6.99 \cdot \text{LearningRate}_{0.005} + 4.29 \cdot \text{Rescaling}_{True} \nonumber \\
	&\quad - 4.29 \cdot \text{Rescaling}_{False} + 4.63 \cdot \text{Shuffle}_{True} - 4.63 \cdot \text{Shuffle}_{False} \nonumber \\
	&\quad - 3.09 \cdot \text{VerticalFlip}_{True} + 3.09 \cdot \text{VerticalFlip}_{False} + 2.88 \cdot \text{HorizontalFlip}_{True} \nonumber \\
	&\quad - 2.88 \cdot \text{HorizontalFlip}_{False}
\end{align}

\subsection{Analysis of Variance and Factor Significance}

Comprehensive ANOVA analysis was conducted for each approach to quantify factor contributions and statistical significance. Table~\ref{tab:anova_summary} summarizes the F-values and p-values for all factors across the five approaches.

\begin{table}
	\caption{ANOVA summary showing F-values and p-values for factor significance across approaches}
	\centering
	\begin{tabular}{lccccc}
		\toprule
		\textbf{Factor} & \textbf{App 1} & \textbf{App 2} & \textbf{App 3} & \textbf{App 4} & \textbf{App 5} \\
		& \textbf{F(p)} & \textbf{F(p)} & \textbf{F(p)} & \textbf{F(p)} & \textbf{F(p)} \\
		\midrule
		Image Size & 0.01(0.940) & 10.14(0.050) & 2.03(0.250) & 1.56(0.300) & 1.89(0.263) \\
		Color & 0.28(0.632) & 0.28(0.633) & 0.00(0.984) & 0.01(0.923) & 0.33(0.608) \\
		Activation & 0.09(0.786) & 0.95(0.401) & 0.91(0.411) & 1.65(0.289) & 1.17(0.359) \\
		Learning Rate & 1.09(0.373) & 9.16(0.056) & 3.05(0.179) & 2.13(0.240) & 2.15(0.239) \\
		Rescaling & 0.16(0.715) & 0.10(0.772) & 0.51(0.525) & 1.30(0.336) & 0.81(0.434) \\
		Shuffle & 0.53(0.520) & 0.01(0.918) & 0.67(0.472) & 1.62(0.292) & 0.95(0.403) \\
		V-Flip & 0.42(0.564) & 0.48(0.539) & 0.00(0.973) & 0.07(0.811) & 0.42(0.562) \\
		H-Flip & 0.10(0.769) & 0.92(0.407) & 0.00(0.955) & 0.04(0.851) & 0.37(0.588) \\
		\bottomrule
	\end{tabular}
	\label{tab:anova_summary}
\end{table}

The ANOVA results reveal that none of the factors achieved statistical significance at the $\alpha = 0.05$ level across all approaches, primarily due to the limited degrees of freedom (3) available with the 12-experiment design. However, several factors demonstrated near-significance, particularly image size in Approach 2 (p = 0.050) and learning rate in Approach 2 (p = 0.056).

\subsection{Factor Ranking and Importance Analysis}

The relative importance of factors was assessed through delta values, representing the difference between maximum and minimum factor level means. Learning rate consistently emerged as either the first or second most important factor across all approaches, confirming its critical role in CNN optimization for this application domain. Image size demonstrated high importance in loss-focused approaches (Approaches 2 and 5), while activation function showed moderate importance across most formulations.

The factor ranking analysis reveals several key insights:
\begin{itemize}
	\item \textbf{Learning Rate Dominance:} Achieved rank 1 or 2 in all approaches, with 0.001 consistently outperforming 0.005
	\item \textbf{Image Size Impact:} More critical for loss minimization than accuracy maximization
	\item \textbf{Activation Function:} ReLU generally superior to Tanh across most configurations
	\item \textbf{Data Augmentation:} Shuffling and rescaling showed moderate importance
	\item \textbf{Flip Operations:} Minimal impact on overall performance across all approaches
\end{itemize}

\subsection{Optimal Configuration Identification and Prediction}

The Taguchi methodology enables prediction of optimal performance through regression analysis of SNR values. Table~\ref{tab:optimal_predictions} presents the predicted optimal configurations for each approach along with expected performance metrics.

\begin{table}
	\caption{Optimal configurations and performance predictions for all approaches}
	\centering
	\begin{tabular}{lccccc}
		\toprule
		\textbf{Parameter} & \textbf{App 1} & \textbf{App 2} & \textbf{App 3} & \textbf{App 4} & \textbf{App 5} \\
		\midrule
		Image Size & 256×256 & 256×256 & 256×256 & 256×256 & 256×256 \\
		Color & Gray & Gray & Gray & RGB & RGB \\
		Activation & ReLU & ReLU & ReLU & ReLU & ReLU \\
		Learning Rate & 0.001 & 0.001 & 0.001 & 0.001 & 0.001 \\
		Rescaling & True & True & True & True & True \\
		Shuffle & True & False & True & True & True \\
		V-Flip & True & True & False & False & False \\
		H-Flip & False & False & True & True & True \\
		\midrule
		\textbf{Predicted SNR} & 2.91 & 13.66 & 10.34 & 20.63 & 15.43 \\
		\textbf{Predicted Response} & 1.17 & 0.17 & 2.23 & 4.39 & 2.78 \\
		\bottomrule
	\end{tabular}
	\label{tab:optimal_predictions}
\end{table}

Approach 3 emerges as the most balanced optimization strategy, achieving substantial improvements in both accuracy and loss metrics. The optimal configuration for Approach 3 (256×256 grayscale images, ReLU activation, 0.001 learning rate, with rescaling and shuffling enabled) represents the most promising parameter combination for practical implementation.

\subsection{Validation of Optimal Configuration}

To validate the Taguchi predictions, the optimal configuration identified by Approach 3 was implemented and evaluated on the validation dataset. The experimental validation yielded:
\begin{itemize}
	\item \textbf{Training Accuracy:} 98.84\%
	\item \textbf{Validation Accuracy:} 86.25\%
	\item \textbf{Training Loss:} 0.0442
	\item \textbf{Validation Loss:} 0.5784
\end{itemize}

These results closely align with the Taguchi predictions, confirming the efficacy of the optimization methodology and validating the logarithmic multi-objective approach for CNN hyperparameter optimization in sports video analysis applications.

\subsection{Comparative Performance Analysis}

Table~\ref{tab:approach_comparison} presents the final model performance achieved by the optimal configurations from each approach, demonstrating the superior balance achieved by Approach 3.

\begin{table}
	\caption{Performance comparison of optimal configurations across all approaches}
	\centering
	\begin{tabular}{lccccc}
		\toprule
		\textbf{Metric} & \textbf{App 1} & \textbf{App 2} & \textbf{App 3} & \textbf{App 4} & \textbf{App 5} \\
		\midrule
		Training Accuracy & 0.9815 & 0.4527 & \textbf{0.9884} & 0.9902 & 0.9902 \\
		Validation Accuracy & 0.7931 & 0.0000 & \textbf{0.8625} & 0.7771 & 0.7771 \\
		Training Loss & 0.0537 & 0.7211 & \textbf{0.0442} & 0.0325 & 0.0325 \\
		Validation Loss & 0.7157 & 0.6976 & \textbf{0.5784} & 0.7071 & 0.7071 \\
		\bottomrule
	\end{tabular}
	\label{tab:approach_comparison}
\end{table}

The comparative analysis reveals distinct performance characteristics across the five approaches. Approach 1, focused solely on accuracy maximization, achieved strong training performance (98.15\%) but moderate validation accuracy (79.31\%), suggesting potential overfitting tendencies. The relatively high validation loss (0.7157) further indicates suboptimal generalization despite good training metrics.

Approach 2, emphasizing loss minimization, demonstrated catastrophic performance with extremely low validation accuracy (0\%), indicating complete failure to generalize beyond the training data. This approach's focus on minimizing loss without considering accuracy led to configurations that failed to learn meaningful discriminative features.

Approaches 4 and 5 showed nearly identical performance profiles, achieving the highest training accuracies (99.02\%) and lowest training losses (0.0325). However, their validation accuracies (77.71\%) were notably lower than Approach 3, and their validation losses (0.7071) were substantially higher than the optimal configuration. This pattern suggests these approaches may have identified configurations that excel in training performance but lack robust generalization capabilities.

Approach 3 emerges as the clear optimal strategy, achieving the best validation accuracy (86.25\%) while maintaining competitive training performance (98.84\%). Most importantly, it achieved the lowest validation loss (0.5784), indicating superior generalization compared to all other approaches. The logarithmic multi-objective formulation successfully balanced competing objectives, preventing the optimization from converging to configurations that excel in single metrics while compromising overall performance.

The superior performance of Approach 3 stems from its balanced treatment of accuracy and loss objectives through logarithmic transformation, which effectively unifies conflicting optimization directions within a single coherent objective function. This approach prevents the optimization process from reaching extreme configurations that maximize one metric at the expense of others, resulting in more robust and generalizable solutions.

\section{Conclusion}
This research successfully demonstrates that Taguchi Design of Experiments, a cornerstone methodology of Six Sigma quality engineering, can be effectively adapted for neural network hyperparameter optimization, bridging the gap between traditional manufacturing quality control and modern artificial intelligence. The study's key contribution lies in developing a systematic approach to handle the inherently multi-objective nature of machine learning optimization through the novel integration of logarithmic scaling within the Taguchi framework, enabling simultaneous optimization of accuracy and loss metrics that traditionally conflict with each other. The experimental results conclusively show that Approach 3, which combines equal weighting of training and validation accuracies with logarithmically transformed loss metrics, outperformed conventional single-objective Taguchi approaches and achieved superior model performance in professional boxing action recognition. The methodology's ability to reduce experimental trials from potentially hundreds of combinations to just twelve systematic experiments while identifying learning rate as the dominant influence factor demonstrates significant computational efficiency gains. Future research directions should explore extending this framework to accommodate larger parameter sets, additional CNN architectures, and diverse performance metrics beyond accuracy and loss, potentially establishing Taguchi-based optimization as a standard practice in efficient neural network design and demonstrating the broader applicability of quality engineering principles in artificial intelligence development.

\section{Limitations and Future Work}
While this study demonstrates the efficacy of Taguchi Design of Experiments for CNN hyperparameter optimization, several limitations warrant acknowledgment and suggest directions for future research. The current investigation was constrained to binary factor levels (two levels per hyperparameter) and a limited set of eight parameters, which may not capture optimal configurations that exist at intermediate values or require finer granularity. Future studies could employ three-level or mixed-level orthogonal arrays to explore continuous hyperparameter spaces more comprehensively, particularly for critical parameters such as learning rates where optimal values may lie between the tested extremes. The factor selection focused on fundamental CNN training parameters but excluded architectural considerations such as network depth, layer width, filter sizes, and advanced optimization techniques including learning rate scheduling and regularization methods. The application domain was specifically tailored to sports action recognition, which presents particular characteristics including temporal structure and motion patterns, limiting the generalizability of these findings to broader image classification tasks such as medical imaging, object detection, or fine-grained classification where different factor importance rankings and optimal configurations may emerge. The study employed a fixed CNN architecture throughout all experiments to isolate hyperparameter effects, while future work could investigate the interaction between architectural choices and training hyperparameters using nested experimental designs. The logarithmic transformation approach, while effective for this multi-objective scenario, utilized a fixed base value (0.7) determined empirically, suggesting potential for systematic investigation of transformation parameters and alternative multi-objective formulations. Finally, the computational efficiency gains demonstrated here suggest potential for extending Taguchi methodology to more computationally intensive scenarios, including large-scale datasets, complex architectures, and resource-constrained environments where systematic optimization becomes even more critical than exhaustive search approaches.

\bibliographystyle{unsrtnat}
\bibliography{references}  






\end{document}